\documentclass[10pt]{article} 

\usepackage[accepted]{rlj} 

\usepackage{amssymb}            
\usepackage{amsthm}             
\usepackage{mathtools}          
\usepackage{mathrsfs}           
\usepackage{graphicx}           
\usepackage{subcaption}         
\usepackage[space]{grffile}     
\usepackage{url}                
\usepackage{lipsum}             
\usepackage{algorithm}
\usepackage{algpseudocode}
\usepackage{amsmath,amsfonts}
\usepackage{hyperref}
\usepackage{microtype}
\usepackage{float}
\usepackage{bm}
\usepackage{caption}
\usepackage{booktabs}
\usepackage{makecell}

\title{Safe Flow Q-Learning: Offline Safe Reinforcement Learning with Reachability-Based Flow Policies}

\setrunningtitle{Safe Flow Q-Learning}

\author{Mumuksh Tayal$^*$\textsuperscript{1}, Manan Tayal$^*$\textsuperscript{1}, Ravi Prakash\textsuperscript{1}}

\emails{mumukshtayal@iisc.ac.in, manantayal@iisc.ac.in, ravipr@iisc.ac.in}

\affiliations{
$^{1}$ \textbf{Indian Institute of Science, India}\\
$^*$\textbf{Denotes equal contribution}\\
}

\contribution{
    We propose \textbf{Safe Flow Q-Learning (SafeFQL)}, a reachability-aware extension of Flow Q-Learning for safe offline reinforcement learning that learns an expressive one-step policy without iterative denoising or rejection sampling at inference.
    }
    {
    The method is evaluated in offline settings with fixed datasets and does not claim online-training safety guarantees.
    }

\contribution{
    We provide a computation-time analysis showing that SafeFQL trades modestly higher offline training cost for substantially lower inference latency than diffusion-style safe generative baselines, supporting real-time deployment in safety-critical loops.
    }
    {
    Latency gains are reported for the evaluated implementations, hardware, and benchmark settings.
    }

\contribution{
    Across boat navigation and all Safety Gymnasium MuJoCo tasks, SafeFQL co-optimizes safety and reward, matching or exceeding prior offline safe RL performance while reducing constraint violations.
    }
    {
    Empirical findings are established on the reported benchmarks and may vary across datasets and task distributions.
    }

\keywords{Safe reinforcement learning, offline reinforcement learning, flow matching, Hamilton-Jacobi reachability.}

\summary{Safe offline reinforcement learning seeks reward-maximizing control from static datasets under strict safety constraints. We propose \textbf{Safe Flow Q-Learning (SafeFQL)}, which extends FQL to the safe setting by combining a Hamilton-Jacobi reachability-inspired safety value function with an efficient one-step flow policy for safe action selection without rejection sampling at deployment. 
Empirically, SafeFQL trades modestly higher offline training cost for substantially lower inference latency than diffusion-style safe generative baselines, making it attractive for real-time safety-critical control. Across boat navigation and all Safe Velocity based Gymnasium MuJoCo tasks, SafeFQL matches or exceeds prior offline safe RL performance while reducing constraint violations.
}

\begin{document}

\makeCover  
\maketitle  

\begin{abstract}
Offline safe reinforcement learning (RL) seeks reward-maximizing policies from static datasets under strict safety constraints. Existing methods often rely on soft expected-cost objectives or iterative generative inference, which can be insufficient for safety-critical real-time control. We propose \textbf{Safe Flow Q-Learning (SafeFQL)}, which extends FQL to safe offline RL by combining a Hamilton--Jacobi reachability-inspired safety value function with an efficient one-step flow policy. SafeFQL learns the safety value via a self-consistency Bellman recursion, trains a flow policy by behavioral cloning, and distills it into a one-step actor for reward-maximizing safe action selection without rejection sampling at deployment. 
Empirically, SafeFQL trades modestly higher offline training cost for substantially lower inference latency than diffusion-style safe generative baselines, which is advantageous for real-time safety-critical deployment. Across boat navigation, and Safety Gymnasium MuJoCo tasks, SafeFQL matches or exceeds prior offline safe RL performance while substantially reducing constraint violations. Project page for the work is available \href{https://tayalmanan28.github.io/safe-fql/}{here}.
\end{abstract}

\section{Introduction}
\label{section: introduction}
Constrained reinforcement learning (CRL) methods incorporate safety objectives during policy learning, but most established approaches rely on extensive online interaction and repeated environment rollouts 
\citep{10.5555/3305381.3305384,altman2021constrained,alshiekh2018safe,Zhao2023SafeRL}. 
This dependence is problematic in safety-critical domains, where training-time failures are costly and many systems do not have sufficiently faithful simulators to absorb risky exploration. As also reflected in safe-RL benchmarks and datasets 
\citep{liu2024dsrl}, the online setting can expose both training and deployment to unacceptable safety risk. 
These limitations motivate a shift toward offline policy synthesis from logged data, including offline RL and imitation-style pipelines 
\citep{levine2020OffRL,kumar2020CQL}. 
However, even in offline settings, many methods still enforce safety through expected cumulative penalties or Lagrangian dual updates, yielding \emph{soft} constraint satisfaction rather than strict state-wise guarantees 
\citep{xu2022constraints,ciftci2024safe,pmlr-v119-stooke20a}. 
Such formulations can be insufficient when a single violation is unacceptable, and the safety-performance trade-off becomes particularly brittle when safety-critical transitions are sparse in static datasets 
\citep{lee2022coptidice}.

Control-theoretic safety methods provide a complementary perspective with stronger notions of state-wise safety. Control Barrier Functions (CBFs) \citep{ames2014control} and Hamilton--Jacobi (HJ) reachability \citep{bansal2017hamilton,Fisac2019HJSafety} can encode forward invariance and worst-case safety explicitly. 
Yet, classical grid-based HJ methods face the curse of dimensionality 
\citep{Mitchell2005ATO}. In addition, many practical CBF/HJ-inspired learning pipelines require either known dynamics or a learned dynamics surrogate to compute safety derivatives and synthesize actions (e.g., through QP-based filtering\citep{ames2017CBF}). 
When dynamics are unknown, model-learning errors can propagate into safety estimates and policy decisions, particularly under dataset shift and out-of-support actions, which weakens practical robustness in purely offline settings 
\citep{tayal2025physics,tayal2025vocbf}. 
Recent offline safety frameworks also report this trade-off explicitly: learned models can enable scalable controller synthesis, but they may become a dominant error source for high-confidence safety if not carefully calibrated 
\citep{tayal2025vocbf}. 
In parallel, safe generative-policy methods have emerged to improve action expressivity under offline distributional constraints. Sequence-model approaches such as the Constrained Decision Transformer (CDT) condition generation on return and cost budgets 
\citep{liu2023constrained}, while diffusion-based methods model multimodal action distributions and can better represent complex behavior support in static datasets
\citep{janner2022planning}.These advances are important because safety-critical datasets are often heterogeneous and multimodal, where unimodal Gaussian actors can fail to recover rare but important safe maneuvers. 
However, current safe generative policies still face practical bottlenecks: sequence-model conditioning is indirect for step-wise safety control, and diffusion-style policies require iterative denoising and often additional rejection sampling to reliably pick safe high-value actions at test time, increasing latency and deployment complexity 
\citep{liu2023constrained,zheng2024fisor}.

At the same time, recent progress in offline RL suggests that improving value learning alone is often insufficient: even with a reasonably accurate critic, extracting an effective policy remains non-trivial 
\citep{park2024BottleneckOffRL}. 
Flow matching provides a useful alternative to diffusion-style generation by learning a continuous transport (velocity-field) map from noise to actions, enabling expressive policy classes with simpler sampling dynamics 
\citep{lipman2023flow}. 
Building on this idea, Flow Q-Learning (FQL) in unconstrained offline RL separates flow-based behavior modeling from one-step RL policy optimization, so the final actor can be optimized efficiently without backpropagating through iterative generation 
\citep{park2025fql}. 
Extending this idea to safety-critical offline RL is not a trivial drop-in adaptation. In the safe setting, policy extraction must simultaneously (i) maximize reward, (ii) remain inside a safety-feasible region under future evolution, and (iii) avoid excessive conservatism that degrades performance.

Motivated by this, we propose \textbf{Safe Flow Q-Learning (SafeFQL)}, an offline safe RL framework that combines reachability-inspired safety value learning with one-step flow policy extraction. SafeFQL learns a safety value function that captures feasibility through a Bellman-style recursion over offline data, and trains a distilled one-step actor that is directly optimized by Q-learning while regularized toward the behavior-supported flow policy. This avoids recursive backpropagation through iterative generative sampling and removes the need for rejection sampling at deployment, while retaining expressive action modeling.

To summarize, our main contributions are:
\begin{itemize}
	\item We formulate \textbf{SafeFQL}, a reachability-aware extension of FQL for safe offline RL that learns an expressive one-step policy without iterative denoising or rejection sampling at inference.
	\item We provide a dedicated \textbf{computation-time analysis} showing that, while SafeFQL may incur higher offline training cost, it delivers substantially lower inference latency than diffusion-style safe generative baselines, enabling real-time deployment in safety-critical control loops.
	\item We show that SafeFQL \textbf{co-optimizes safety and performance} across custom navigation and Safety Gymnasium benchmarks, consistently achieving lower safety violations while maintaining strong reward relative to prior constrained offline RL and safe generative baselines.
\end{itemize}


\section{Background and Problem Setup}
\label{section: background}
We study safe offline reinforcement learning in environments with hard state constraints. The environment is modelled as a Constrained Markov Decision Process (CMDP), defined by the tuple
$\mathcal{M} = (\mathcal{X}, \mathcal{A}, P, r, \ell, \gamma)$,
where $\mathcal{X}$ and $\mathcal{A}$ denote the state and action spaces, $P(x'|x,a)$ denotes the transition probability function defining the system dynamics, $r : \mathcal{X} \rightarrow \mathbb{R}$ is the reward function, $\ell : \mathcal{X} \rightarrow \mathbb{R}$ is an instantaneous state-based safety function, typically defined as the negative of signed distance function to failure set $\mathcal{F}$, and $\gamma \in (0,1)$ is the discount factor. We define the \emph{failure set} $\mathcal{F} := \{x \in \mathcal{X} \mid \ell(x) > 0\}$, which represents unsafe states that must be avoided at all times (e.g., collisions or constraint violations). A trajectory is considered safe if it never enters $\mathcal{F}$. We assume access to an offline dataset $\mathcal{D} = \{(x_t, a_t, r_t, \ell_t, x_{t+1})\}$, collected by an unknown behavior policy, with no further interaction with the environment permitted. Any policy $\pi(a \mid x)$ which induces trajectories $\Gamma = (x_0, a_0, x_1, \dots)$, does it through the transition probability function $P$.

Given an initial state $x$, the objective is to compute the maximum achievable discounted return subject to state safety at all future time steps. This requirement can be formalized through the following formulation:
\begin{equation}
\label{eq:scorl}
\begin{aligned}
    \sup_{\pi}
\big[ \sum_{k=0}^{\infty} \gamma^k r(x_k) \,\big|\, x_0 = x \big]
\;\;\\  \text{s.t.} \;\;
x_t \notin \mathcal{F}, \;\forall t \ge 0.
\end{aligned}
\end{equation}

Unlike formulations based on expected cumulative penalties, \eqref{eq:scorl} encodes a \emph{hard safety requirement}, i.e., only policies that admit trajectories remaining entirely outside the failure set are considered feasible. This formulation directly captures safety-critical requirements where even a single violation is unacceptable.

\subsection{Generative Policies for Offline RL}
To overcome the limitations of traditional limitations for policy extraction, recent literature has investigated generative policy representations in offline RL, such as sequence models and diffusion-based policies \citep{chen2021decision,janner2022planning}, along with their extensions to safety-constrained environments \citep{liu2023constrained,lin2023safe,zheng2024fisor, liu2025ciql}. Although highly effective at capturing data distributions, diffusion models necessitate the simulation of stochastic processes across numerous discrete time steps during inference. This iterative sampling is computationally burdensome, making real-time deployment in high-frequency control loops particularly challenging. Conversely, flow matching \citep{lipman2023flow, zhang2025EWFM, alles2025flowq} presents a deterministic alternative. By directly learning the vector field of the generative process, flow matching facilitates highly efficient policy sampling through a single ODE integration. A convenient way to view flow-matching policies is as the time-1 pushforward of a state-conditioned, time-dependent velocity field. Let \(v_{\theta}(t,x,z)\) denote the state-conditioned velocity field and define the flow \(\psi_{\theta}(t,x,z)\) by the ODE
\begin{align}
\frac{d}{dt}\,\psi_{\theta}(t,x,z)=v_{\theta}\big(t,x,\psi_{\theta}(t,x,z)\big),\qquad
\psi_{\theta}(0,x,z)=z.
\end{align}
The corresponding flow policy is defined as the ODE terminal map
\begin{align}
\mu_{\theta}(x,z)\coloneqq\psi_{\theta}(1,x,z)
= z + \int_{0}^{1} v_{\theta}\big(t,x,\psi_{\theta}(t,x,z)\big)\,dt,
\end{align}
which is a deterministic mapping in $(x,z)$ but induces a stochastic policy \(\pi_{\theta}(a\mid x)\) via \(z\sim\mathcal{N}(0,I)\). We will dive deeper into this aspect of deterministic mapping of $(x,z)$ in the later sections.

\subsection{Safe Offline Reinforcement Learning}
Safe reinforcement learning has conventionally relied on online Lagrangian-based constrained optimization and trust-region methods \citep{chow2017algorithm,tessler2018reward,pmlr-v119-stooke20a,10.5555/3305381.3305384}. However, the necessity for online interaction and the use of soft cost penalties in these approaches have catalyzed a shift toward safe offline RL. Several prominent offline RL methods such as CPQ \citep{xu2022constraints} and C2IQL \citep{liu2025ciql} attempt to ensure safety by penalizing unsafe actions by restricting the expected cumulative costs below a pre-defined cost limit $l$, i.e., $\max_{\pi} \mathbb{E}_{\tau\sim\pi}[\sum_{t=0}^{\infty}\gamma^t r(x_t,a_t)] \quad \text{s.t.}~\mathbb{E}_{\tau\sim\pi}[\sum_{t=0}^{\infty}\gamma^t c(x_t)] \le l$; but these techniques often degrade value estimation and generalization \citep{li2023when}. 

Some Hamilton--Jacobi (HJ) reachability based safety frameworks connect HJ reachability with offline RL \citep{zheng2024fisor} to identify states that can enter the failure set \(\mathcal{F}=\{x:\ell(x)\ge0\}\) within a given time horizon \citep{bansal2017hamilton,Fisac2019HJSafety}. They often define the HJ value as the best worst-time safety margin
\begin{align}
&V_{\ell}^*(x_0)\;\coloneqq\;\inf_{\pi}\sup_{t\in[0,T]} \ell(x_t)
\quad\text{s.t.}~a_t\sim\pi(\cdot\mid x_t),\\
\text{Or,} \quad &V_{\ell}^*(x_0)\;\coloneqq\;\max\{\ell(x_0), \inf_{\pi}V_{\ell}^{\pi}(x_1)\} \quad \forall~t\in\{0, 1, 2, ...\}
\end{align}
so that $V_{\ell}^*(x_0)$ measures the smallest maximum value of \(\ell\) attainable along trajectories from $x_0$.  Intuitively, $V_{\ell}^*(x_0)>0$ indicates that even the best policy leads the trajectory inside the failure set (i.e., $\ell(x_t)\ge0$ for some $t$), while $V_{\ell}^*(x_0)<0$ implies there exists an optimally safe policy that keeps the system in the safe region from state $x_0$ within the horizon. The classical HJ PDE / Hamiltonian formulation and numerical solution methods are used for computation \citep{bansal2017hamilton}. Such frameworks often use Generative Policy based techniques like DDPM \citep{zheng2024fisor} and Flow Matching to learn expressive policies in offline RL. However, such frameworks struggle to extract an exact optimal policy and rather tend to learn a policy which only encourages the desired safety and performance with the use of Advantage Weighted Regression \citep{Peters2007RLwithAWR}. And even though AWR is a simple and easy-to-implement approach in offline RL, it is often considered as the least effective policy extraction method \citep{park2024BottleneckOffRL}, and therefore, many a times has to be accompanied by Rejection Sampling to selectively choose an action that best suites the requirements. Perhaps, a more effective technique for policy extraction can be using Deterministic Policy Gradient with Behavior Cloning \citep{fujimoto2021ddpg} where the policy directly maximizes Q-value function. But using DPG with multi-step denoising based generative policy frameworks like Flow Matching requires backward gradient through the entire reverse denoising process, inevitably introducing tremendous computational costs. Meanwhile, other set of frameworks use Barrier Function based approaches \citep{wang2023enforcing, tayal2025vocbf} to achieve safety. Unfortunately, these frameworks also come with their own set of limitations. Barrier Functions require knowledge of system dynamics which is generally rare to be known for most systems. Although such frameworks choose to learn the approximate dynamics of the system, they can become a significant source of noise, which can be fatal in safety-critical cases.

To overcome these bottlenecks, recent works have focused on distilling the multi-step generative processes into single-step policies \citep{prasad2024consistencypolicy, zhang2025DPD, park2025fql}. These distilled models are designed to match the action outputs of their full-fledged, multi-step counterparts, yielding fast and accurate performance at a fraction of the computational cost for both training and inference.

\section{Safe Flow Q-Learning}
\label{section: method}
\begin{figure}[htb]
    \centering
    \includegraphics[width=\linewidth]{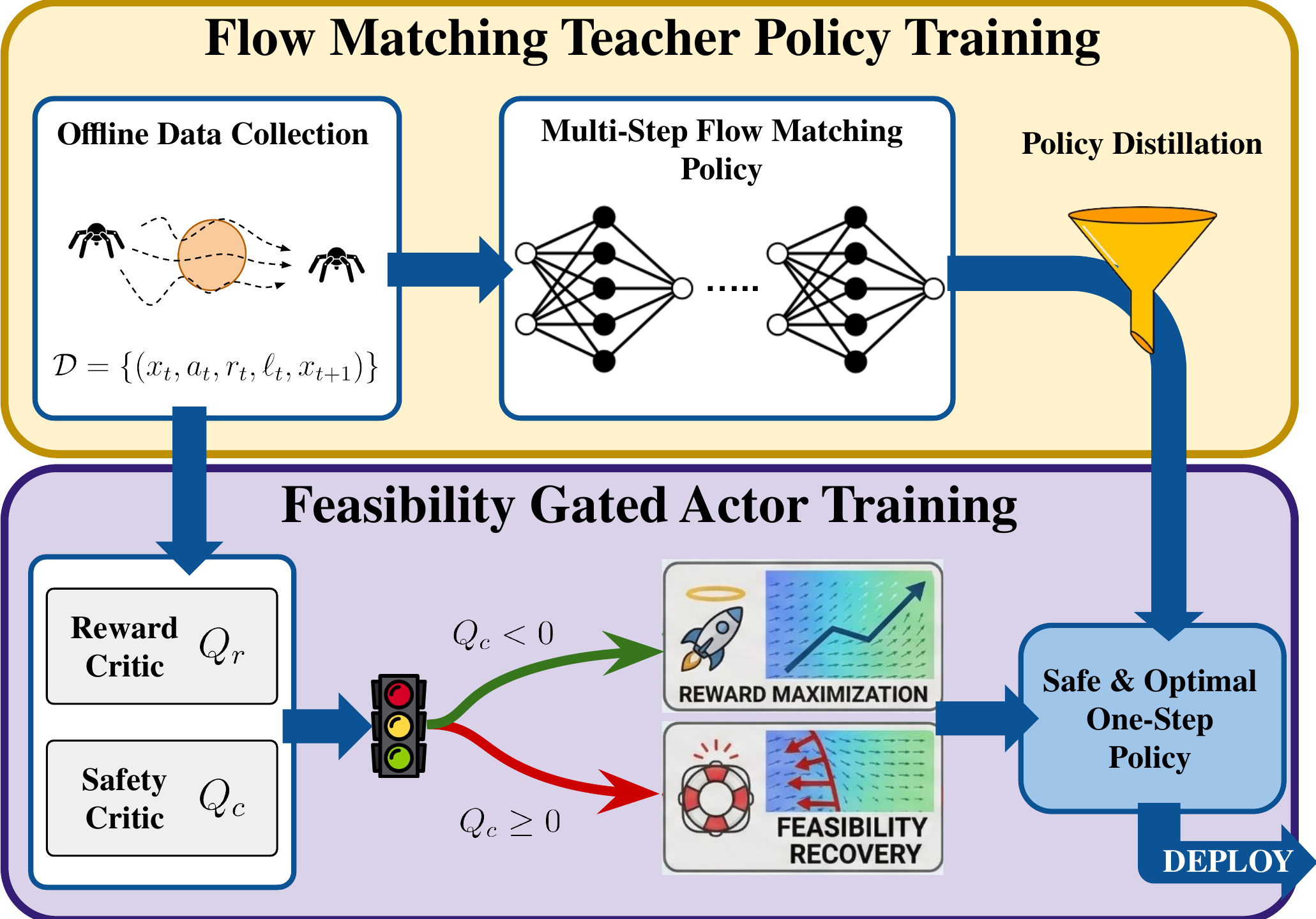}
    \caption{\textbf{Framework Overview.} SafeFQL framework proposes a 2 stage safe offline RL approach which uses an efficient one-step flow policy extraction using policy distillation and feasibility gating.}
    \label{fig:placeholder}
\end{figure}

Building on the CMDP formulation and the offline safe RL objective introduced in Section~\ref{section: background}, this section presents SafeFQL in full detail. The design follows the decoupled learning principle of FQL \citep{park2025fql} where value functions and the policy are trained with separate objectives so that policy optimization is never destabilized by errors in critic bootstrapping. We extend this principle to the safety-constrained setting by introducing a second critic system whose semantics are governed by worst-case reachability rather than cumulative discounted return. 

The overall procedure decomposes into three phases: \emph{(i)} learning reward and safety critics from $\mathcal D$; \emph{(ii)} fitting a behavior flow teacher and distilling it into a one-step actor; and \emph{(iii)} optimizing the actor under a feasibility-gated objective.
These three phases are sequentially dependent, the policy cannot be trained before critics converge. Within each phase, all networks are trained in parallel to convergence. We describe each phase in turn.

\subsection{Learning Reward and Safety Critics}
\label{sec:critics}

The offline dataset $\mathcal D=\{(x_i,a_i,r_i,\ell_i,x_i')\}_{i=1}^N$ provides tuples of state, action, scalar reward, signed safety signal, and next state. We recall from Section~\ref{section: background} that the safety signal $\ell(x)$ is defined so that $\ell(x)\le 0$ if and only if $x\notin\mathcal F$, i.e., the state is safe. All critic learning is performed entirely within the support of $\mathcal D$, so that no out-of-distribution action queries are required.

\paragraph{Reward critics.}
We train a reward Q-function $Q_r(x,a;\phi_r)$ and a corresponding state-value function $V_r(x;\psi_r)$ using the implicit Q-learning (IQL) approach of \citet{kostrikov2022offline}. IQL avoids querying the actor during critic updates, which is the primary source of instability in offline actor--critic methods \citep{fujimoto2019off}. The value function $V_r$ approximates the expectile of the Q-value distribution under the behavior policy, and is trained via the asymmetric squared loss
\begin{align}
\mathcal L_{V_r}(\psi_r)
=\mathbb E_{(x,a)\sim\mathcal D}
\!\left[\mathcal L_\tau\!\left(Q_r(x,a;\phi_r)-V_r(x;\psi_r)\right)\right],
\label{eq:vr_loss}
\end{align}
where $\mathcal L_\tau(u)=|\tau-\mathbb I(u<0)|\,u^2$ is the expectile loss with $\tau\in(0.5,1)$. For $\tau$ close to 1 the loss upweights positive residuals, causing $V_r$ to track a high quantile of the in-sample Q-value distribution rather than its mean. This implicitly represents the advantage of actions better than average in the dataset without ever evaluating the policy. Given $V_r$, the Q-function is updated via one-step Bellman regression against a target network $\bar V_r$:
\begin{align}
y_r &= r + \gamma\,\bar V_r(x'), \label{eq:qr_target}\\
\mathcal L_{Q_r}(\phi_r)
&= \mathbb E_{(x,a,r,x')\sim\mathcal D}\!\left[\left(Q_r(x,a;\phi_r)-y_r\right)^2\right].
\label{eq:qr_loss}
\end{align}
Target network parameters $\bar\psi_r$ are updated via Exponential Moving Average (EMA), details for which are covered in Supplementary Material~\ref{appendix: exp_details}.

\paragraph{Safety critics.}
For the safety constraint, a naive approach would be to train a discounted cumulative cost Q-function $Q_c^{\text{sum}}(x,a)=\mathbb E[\sum_t \gamma^t \mathbb I\{x_t\in\mathcal F\}]$ and penalize its expectation below a threshold, as in standard CMDP Lagrangian methods. This leads to a soft constraint that enforces safety in expectation but cannot prevent individual trajectory violations \citep{xu2022constraints,lee2022coptidice}. Moreover, the non-negativity of the cumulative cost makes the threshold a free hyperparameter that must be tuned per task.

SafeFQL instead adopts a reachability-inspired formulation that encodes \emph{worst-case} safety along the trajectory. We define the safety critic $Q_c(x,a)$ as an approximation of the Hamilton--Jacobi feasibility value $V_\ell^*(x_0)=\min_\pi\max_{t\ge 0}\ell(x_t)$ from Section~\ref{section: background}, trained via a \emph{max-backup} Bellman recursion \citep{Fisac2019HJSafety}:
\begin{align}
y_c(x,a,x') = \max\!\left\{\ell(x),\;\gamma\,\bar V_c(x')\right\}.
\label{eq:yc}
\end{align}
The target $y_c$ takes the \emph{maximum} of the immediate safety margin $\ell(x)$ and the discounted future safety value $\gamma\bar V_c(x')$. This ensures that a low safety margin at \emph{any} future time step propagates backward to the current state, so that $Q_c(x,a)< 0$ carries a strong meaning: not only is $x$ currently safe, but the predicted future evolution also remains in the safe region under behavior-policy-like actions. Conversely, $Q_c(x,a)\ge0$ indicates that following the behavior distribution from $(x,a)$ is predicted to eventually enter the failure set $\mathcal F$.

The safety Q-function $Q_c(x,a;\phi_c)$ and safety value function $V_c(x;\psi_c)$ are trained with
\begin{align}
\mathcal L_{Q_c}(\phi_c)
&= \mathbb E_{(x,a,\ell,x')\sim\mathcal D}\!\left[\left(Q_c(x,a;\phi_c)-y_c\right)^2\right],
\label{eq:qc_loss}\\[4pt]
\mathcal L_{V_c}(\psi_c)
&= \mathbb E_{(x,a)\sim\mathcal D}
\!\left[\mathcal L_\tau\!\left(Q_c(x,a;\phi_c)-V_c(x;\psi_c)\right)\right].
\label{eq:vc_loss}
\end{align}
Note the use of the same expectile loss in \eqref{eq:vc_loss} and \eqref{eq:vr_loss}, but applied to the safety residual $Q_c - V_c$. Here $\tau<0.5$ causes $V_c$ to track the \emph{lower} quantile of the in-sample safety Q-distribution, yielding a conservative approximation of the feasibility boundary. In implementation we share the same $\tau$ hyperparameter across both critics, with opposite sign conventions in expectile regression (i.e., $u>0$ where $u = Q_c(x,a;\phi_c) - V_c(x;\psi_c)$) for what constitutes a desirable extreme; the reward critic targets the upper quantile while the safety critic targets the lower quantile. The max-backup structure of $y_c$ means that clipped double-Q techniques familiar from reward critics must be applied with a \emph{maximum} operation (i.e., taking the most pessimistic safety estimate): we use two safety Q-networks and set $\bar V_c(x')=\max\{Q_c^{(1)}(x',\cdot), Q_c^{(2)}(x',\cdot)\}$, consistently avoiding overoptimistic feasibility estimates at OOD next states.

\subsection{Behavior Flow Policy and One-Step Distillation}
\label{sec:flow}

With critics in place, we turn to policy learning. The central challenge is to produce a policy that (a) stays close to the behavior distribution to avoid distributional shift, (b) is expressive enough to model multimodal and structured action distributions common in robotics datasets, and (c) can be executed at test time with negligible latency. Diffusion-based policies satisfy (a) and (b) through score-matched generative modeling, but their iterative reverse-process sampling incurs $O(T)$ network evaluations per step \citep{zheng2024fisor,11}. SafeFQL therefore adopts the FQL strategy \citep{park2025fql} of using a flow-matching model as a \emph{fixed behavior teacher} and distilling it into an efficient one-step deployment policy.

\paragraph{Flow behavior teacher.}
We parameterize the behavior policy $\pi_\beta$ via a conditional flow-matching model $\mu_\theta(x,z,t)$, which defines a time-dependent velocity field over actions \citep{lipman2023flow}. Given a state $x\sim\mathcal D$, a Gaussian sample $z\sim\mathcal N(0,I)$, and a time $t\sim\text{Uniform}([0,1])$, the teacher is trained to transport $z$ to the empirical action distribution via the regression objective
\begin{align}
\mathcal L_{\text{flow}}(\theta)
=\mathbb E_{(x,a)\sim\mathcal D,\,z\sim\mathcal N(0,I),\,t\sim\mathcal U([0,1])}
\!\left[\left\|\mu_\theta(x,\,x_t,\,t) - (a - z)\right\|_2^2\right],
\label{eq:flow_teacher}
\end{align}
where $x_t=(1-t)z+ta$ is the straight-line interpolation between the noise sample and the target action. At convergence, integrating the learned velocity field from $t=0$ to $t=1$ starting from $z$ generates an action $a\sim\pi_\beta(\cdot|x)$. The flow teacher is trained with behavioral cloning only (no critic signal enters $\mathcal L_{\text{flow}}$), which keeps this stage unconditionally stable.

\paragraph{One-step student actor.}
The deployed policy is a deterministic one-step actor $\mu_\omega(x,z):\mathcal X\times\mathbb R^{d_a}\to\mathcal A$ that maps a state and a latent noise vector directly to an action, without any iterative integration. To endow the student with the expressiveness of the flow teacher, we define a \emph{distillation loss} that penalizes deviation from the one-step teacher output $\tilde\mu_\theta(x,z)$, which is the action produced by running a single integration step of the trained flow model from $z$ conditioned on $x$:
\begin{align}
\mathcal L_{\text{distill}}(\omega)
=\mathbb E_{(x,z)\sim\mathcal D\times\mathcal N(0,I)}
\!\left[\left\|\mu_\omega(x,z)-\tilde\mu_\theta(x,z)\right\|_2^2\right].
\label{eq:distill}
\end{align}
The distillation term serves as a behavior regularizer: it pulls the student actor toward the support of the offline dataset, preventing it from exploiting critic extrapolation errors in regions far from the data \citep{park2025fql}. Crucially, the one-step actor $\mu_\omega$ is the \emph{only} component of SafeFQL that is queried at deployment time, so inference cost is that of a single forward pass through a feedforward network regardless of how many flow steps were used to train the teacher.

\subsection{Feasibility-Gated Actor Objective}
\label{sec:actor}

Given the reward critic $Q_r$, the safety critic $Q_c$, and the distillation anchoring from the flow teacher, we now describe how to combine these signals into a well-motivated actor objective. This is the crux of the method, and the design choice here distinguishes SafeFQL from both vanilla FQL and prior soft-constraint offline safe RL approaches.

\paragraph{Limitations of the naive Lagrangian formulation.}
A natural baseline is to treat the safety constraint as a soft penalty and jointly optimize reward and safety with a Lagrangian multiplier $\eta>0$:
\begin{align}
\mathcal L_{\text{actor}}^{\text{naive}}(\omega)
= \mathbb E_{x,z}\!\left[-Q_r(x,a_\omega)+\eta\,\max\!\left(0,Q_c(x,a_\omega)\right)\right]
+\lambda\,\mathcal L_{\text{distill}}(\omega),
\label{eq:naive_lag_actor}
\end{align}
where $a_\omega=\mu_\omega(x,z)$. The penalty term $\max(0,Q_c)$ is zero when $Q_c<0$ (predicted feasible) and equal to $Q_c$ when $Q_c\ge0$ (predicted infeasible). Objective~\eqref{eq:naive_lag_actor} is computationally straightforward since gradients with respect to $\omega$ flow through both terms simultaneously, but it has a critical structural flaw, the two terms are commensurate in magnitude and can trade off against each other. Concretely, near the feasibility boundary where $Q_c$ is small but positive, a sufficiently large gradient from $-Q_r$ can dominate and push the actor into the infeasible region. The multiplier $\eta$ would need to be tuned precisely per task to prevent this, and the right value is not known without online interaction. Empirically, soft-constraint offline methods that rely on this kind of Lagrangian penalty are known to be highly sensitive to the choice of cost limit and multiplier \citep{zheng2024fisor,xu2022constraints}; the coupled optimization of reward, safety, and behavior regularization further exacerbates instability \citep{lee2022coptidice}.

\paragraph{Feasibility-gated objective.}
SafeFQL replaces the additive tradeoff with an \emph{exclusive-gate} mechanism that enforces strict priority ordering, when the predicted policy action violates feasibility ($Q_c\ge0$), the actor update completely ignores reward and focuses solely on recovering feasibility; only once the action is predicted feasible ($Q_c<0$) does the update switch to reward maximization. This is implemented via the binary gate
\begin{align}
\zeta(x,z) = \mathbb I\!\left\{Q_c\!\left(x,\mu_\omega(x,z)\right)< 0\right\},
\label{eq:gate}
\end{align}
and the combined actor loss
\begin{align}
\mathcal L_{\text{actor}}(\omega)
= \lambda\,\mathcal L_{\text{distill}}(\omega)
+\mathbb E_{(x,z)}\!\left[
  \zeta(x,z)\cdot\big(-Q_r(x,a_\omega)\big)
  +\big(1-\zeta(x,z)\big)\cdot\max\!\left(0,Q_c(x,a_\omega)\right)
\right].
\label{eq:safefql_actor}
\end{align}
The three terms in \eqref{eq:safefql_actor} have distinct and complementary roles. The distillation term $\mathcal L_{\text{distill}}$ serves as a universal behavioral anchor, keeping the actor within the support of the offline dataset at all times regardless of the feasibility state. The second term $\zeta\cdot(-Q_r)$ is the reward-maximization signal, which is active only at state-latent pairs where the current actor output is already in the predicted feasible region. The third term $(1-\zeta)\cdot\max(0,Q_c)$ is the feasibility recovery signal, active only when the current output violates the predicted safety boundary, and it pushes the actor in the direction of decreasing $Q_c$ rather than in the direction of reward. Such a feasibility gate $\zeta$ eliminates the instability that arises when reward and safety gradients are simultaneously active and point in opposing directions.


\paragraph{In-sample action generation.}
At each actor update step, the action $a_\omega=\mu_\omega(x,z)$ with $z\sim\mathcal N(0,I)$ is sampled fresh, so the stochastic policy $\pi_\omega$ induced by $\mu_\omega$ is implicitly evaluated at many points per gradient step. No replay buffer of policy actions is needed; the randomness of $z$ provides the necessary coverage of the action distribution to avoid mode collapse under the distillation constraint.

\begin{algorithm}[t]
\caption{Safe Flow Q-Learning (SafeFQL)}
\label{algo:safe-fql}
\begin{algorithmic}[1]
\Require Offline dataset $\mathcal D=\{(x,a,r,\ell,x')\}$, discounts $\gamma$, expectile $\tau$, distillation weight $\lambda$, calibration parameters $(\epsilon_s,\beta_s,N_{\mathrm{cal}})$
\Ensure Deployed policy $\mu_\omega$; corrected safe set $\mathcal S_{\delta^*}$
\vspace{2pt}
\State \textsc{// Phase 1: Critic Learning}
\State Initialize $Q_r,V_r$ (reward critics) and $Q_c,V_c$ (safety critics)
\For{each gradient step}
  \State Update $V_r$ via expectile loss~\eqref{eq:vr_loss}; update $Q_r$ via Bellman loss~\eqref{eq:qr_loss}
  \State Update $Q_c$ via max-backup Bellman loss~\eqref{eq:qc_loss}; update $V_c$ via expectile loss~\eqref{eq:vc_loss}
  \State EMA-update target networks $\bar V_r$, $\bar V_c$
\EndFor
\vspace{2pt}
\State \textsc{// Phase 2: Flow Teacher Training}
\State Train behavior flow teacher $\mu_\theta$ via flow-matching loss~\eqref{eq:flow_teacher}
\vspace{2pt}
\State \textsc{// Phase 3: Feasibility-Gated Actor Training}
\State Initialize one-step actor $\mu_\omega$
\For{each gradient step}
  \State Sample $(x,z)\sim\mathcal D\times\mathcal N(0,I)$; compute $a_\omega=\mu_\omega(x,z)$
  \State Compute gate $\zeta$ via~\eqref{eq:gate}; update $\mu_\omega$ by minimizing~\eqref{eq:safefql_actor}
\EndFor
\vspace{2pt}
\State \Return $\mu_\omega$
\end{algorithmic}
\end{algorithm}

\section{Experiments}
\label{section: experiments}
We evaluate SafeFQL against baseline methods to investigate three critical aspects: (i) its safety rate relative to state-of-the-art constrained offline RL algorithms, (ii) the tradeoff between safety compliance and cumulative reward $\sum_{k=0}^\infty r(x_k)$, and (iii) its sampling efficiency during inference compared to prominent alternative generative modeling based frameworks. Our findings confirm that SafeFQL successfully co-optimizes safety and performance, attaining higher safety rates while maintaining competitive reward accumulation.

\textbf{Baselines:} We compare SafeFQL against a diverse set of safety constrained offline reinforcement learning methods. We include \textbf{BEAR-Lag} (Lagrangian dual version of \citet{kumar2019stabilizing}), \textbf{COptiDICE} \citep{lee2022coptidice}, \textbf{CPQ} \citep{xu2022constraints}, \textbf{C2IQL} \citep{liu2025ciql}, \textbf{FISOR} \citep{zheng2024fisor} and also \textbf{SafeIFQL} (Safe Flow Matching version of \citet{kostrikov2022offline}). In contrast to these methods, SafeFQL learns a one step policy from offline demonstrations that accounts for future unsafe interactions in advance and therefore, accordingly taking actions that maximize the cumulative reward while staying within the safe region.

\nocite{tayal2026epiflow}

\textbf{Evaluation Metrics:} We evaluate all methods based on (i) \emph{safety/cost}, measured as the total number of safety violations incurred before episode termination, and (ii) \emph{performance}, measured via the cumulative episode rewards. These metrics allow us to assess the trade-off between strict safety enforcement and task performance across different offline RL approaches.

\subsection{Experimental Case Studies}

\begin{figure}[htb]
  \centering
  \includegraphics[width=0.99\linewidth]{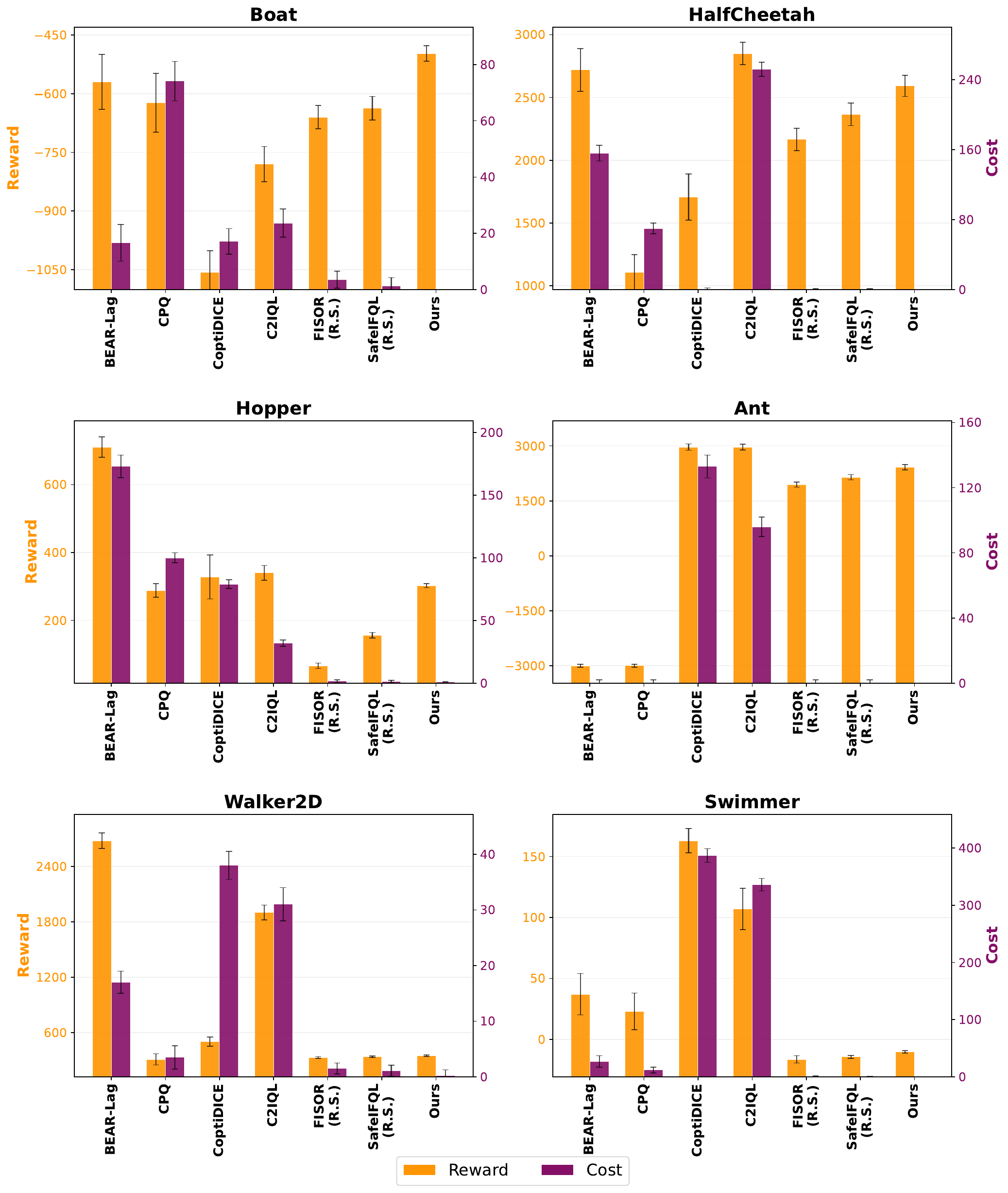}
  \caption{\textbf{Evaluation Results.} Reports reward and cost (mean $\pm$ std dev) in all evaluation environments for all frameworks. SafeFQL achieves the lowest costs across all the evaluated environments while achieving highest reward among the frameworks with comparable costs. Some baselines with \emph{(R.S.)} tag represent frameworks that are evaluated using Rejection Sampling (N=16) at evaluation time. Results evaluated across 5 seeds.}
  \label{fig:results}
\end{figure}

For thorough evaluation of our framework against the baselines, we use the following varied set of environments:
\begin{itemize}
    \item \textbf{Safe Boat Navigation:} For our first experiment, we address a two-dimensional collision avoidance problem where a boat, modeled with point mass dynamics, navigates a river \citep{tayal2025vocbf}. The river's drift velocity changes according to the boat's y-coordinate. The primary goal is to safely bypass obstacles despite this variable drift. We evaluated our approach against baseline methods using fixed 500 randomly selected initial states. Comprehensive details regarding the system dynamics, state space boundaries, and experimental setup can be found in Supplementary Material~\ref{appendix: Boat2D}.
    
    \item \textbf{Safety Gymnasium:} To further validate our framework, we conducted evaluations within Safety Gymnasium \citep{ji2023safety}, specifically focusing on the Safe Velocity suite. For these Safe Velocity tasks, we tested SafeFQL on several high-dimensional MuJoCo environments, including Hopper, Half Cheetah, Swimmer, Walker2D, and Ant. The goal in these settings is to maximize the agent's reward while strictly maintaining its speed below a specified threshold. To ensure a fair comparison against baseline methods, we evaluated performance across a fixed set of 500 randomly sampled initial states for each environment. All the additional details of the framework are available in Supplementary Material~\ref{appendix: SafetyGym} for the readers to refer.
\end{itemize}

For our experiments within the Safety Gymnasium suite, we employ the standard DSRL dataset for safe offline RL \citep{liu2024dsrl} while preserving the framework's original reward and safety-violation metrics \citep{ji2023safety}. To benchmark our approach against existing baselines, we evaluate performance across a fixed collection of 500 randomly sampled safe initial states for each task.

\subsection{Results}
While refering the results from Figure~\ref{fig:results} for the custom Safe Boat Navigation environment, SafeFQL achieves a significant increase in reward compared to all baselines while maintaining zero violations across all evaluation episodes . This strong performance extends to the high-dimensional Safety Gymnasium tasks (HalfCheetah, Hopper, Ant, Walker2D, and Swimmer), where SafeFQL consistently achieves the lowest safety violations and the highest reward among frameworks with comparable near-zero costs . SafeFQL's success stems from learning a one-step optimal policy that directly outputs high-reward, safe actions . In contrast, baselines optimizing for expected cumulative cost (e.g., BEAR-Lag, CPQ, and COptiDICE) struggle to strictly enforce safety without sacrificing reward , while C2IQL achieves high rewards but incurs inconsistent costs in safety-critical task.

Furthermore, while generative baselines like FISOR and SafeIFQL also account for worst-case safety, they fail to directly learn a single optimal action. Instead, they rely on computationally expensive and suboptimal rejection sampling at inference to filter safe actions from multiple generated candidates. SafeFQL circumvents this entirely by directly outputting the optimal action at each timestep without the need for sampling, establishing it as the most effective and efficient policy among the evaluated frameworks .

\subsection{Comparison to Generative Policy Baselines}
\begin{figure}[ht]
  \centering
  \includegraphics[width=0.8\linewidth]{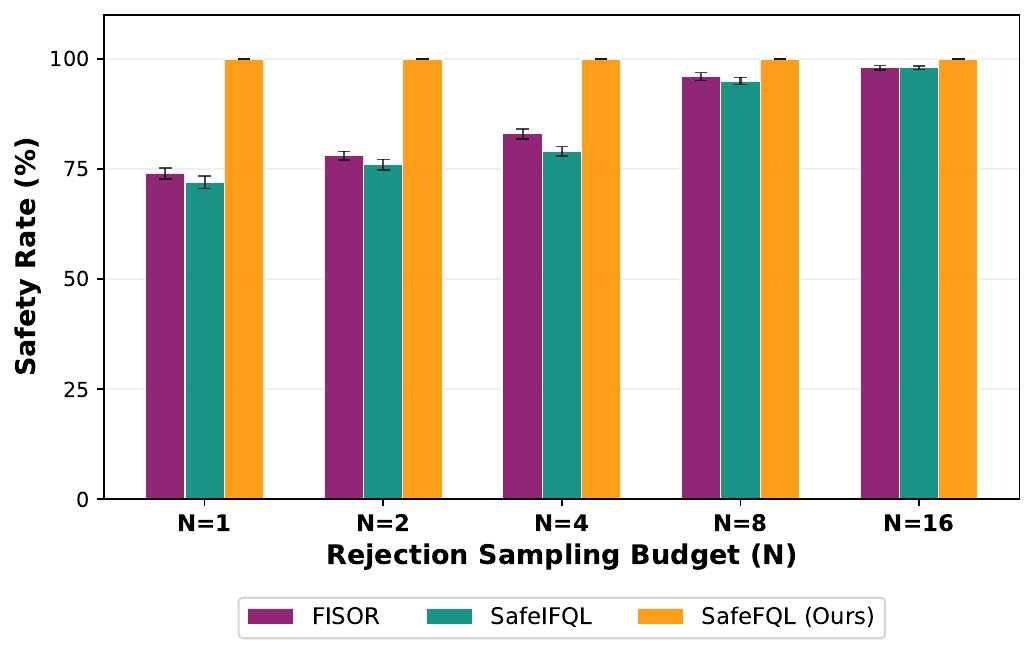}
  \caption{\textbf{Action Sampling Efficiency.} Generative policy–based methods (FISOR, SafeIFQL) require rejection sampling to reach high safety (mean $\pm$ std dev) across 5 seeds; SafeFQL achieves highest safety in the Safe Boat Navigation environment with only N=1 action sample, while other baselines require larger N.}
  \label{fig:boat-safety-rate}
\end{figure}

To further investigate the performance benefits of SafeFQL, we analyzed the safety rate, defined as the percentage of episodes without collisions, in the Figure~\ref{fig:boat-safety-rate} for the Safe Boat Navigation environment. Generative baseline frameworks like FISOR and SafeIFQL require a large number of action samples (N=16) to achieve safety rates comparable to our method. Because SafeFQL learns a one-shot optimal policy, it successfully outputs the optimal action even when N is restricted to 1.

\begin{figure}
    \centering
    \includegraphics[width=0.96\linewidth]{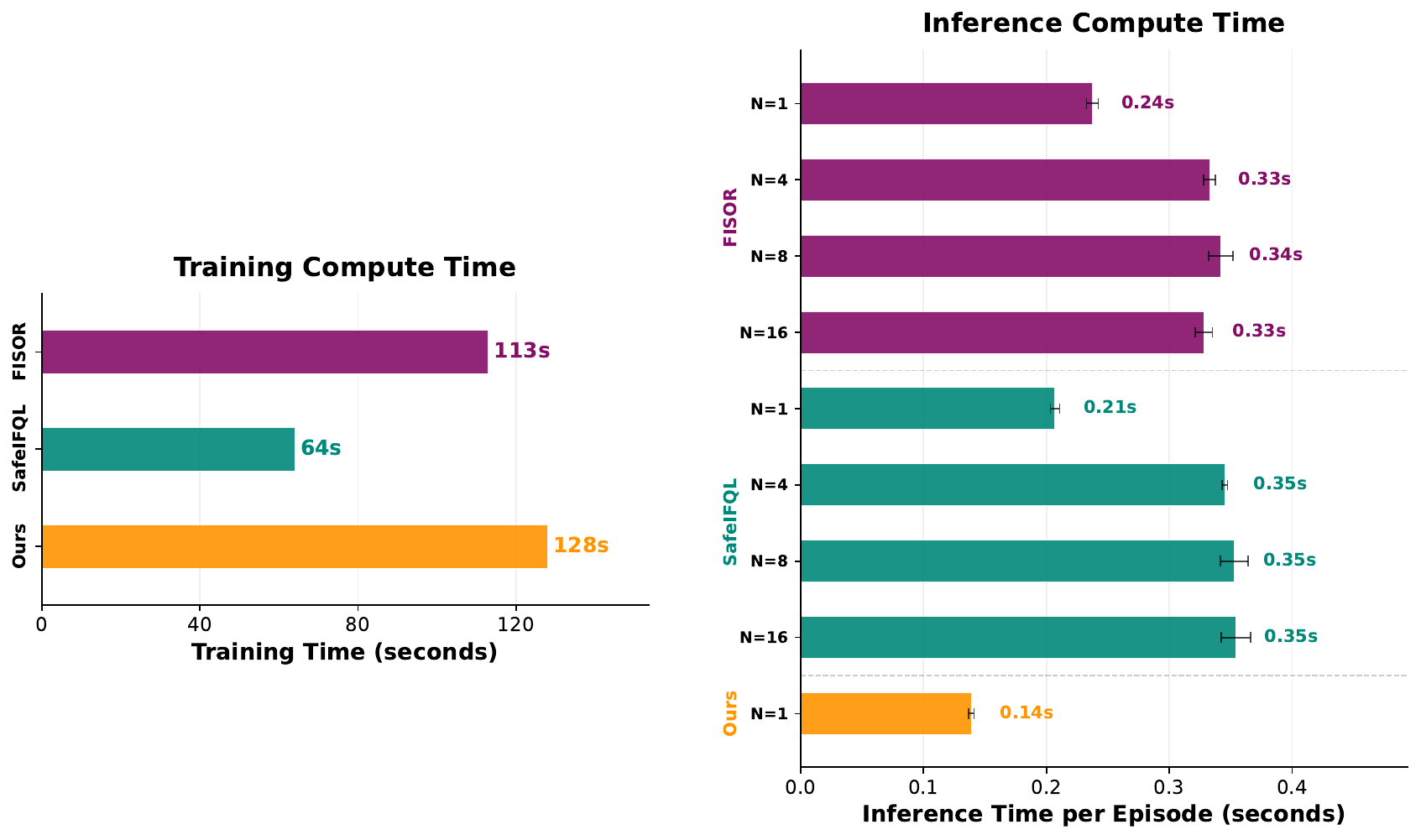}
    \caption{\textbf{Computation Time Analysis.} Training Time \emph{(Left)} and Inference Time \emph{(Right)} (mean $\pm$ std dev) taken by each of the three generative policy based frameworks when evaluated across 5 independent trial runs.}
    \label{fig:compute-time}
\end{figure}

Besides, we also evaluated the computation time gains of SafeFQL over FISOR and SafeIFQL by analyzing both training and inference times in Figure~\ref{fig:compute-time}. While SafeFQL requires a longer training compute time compared to the other two frameworks, it more than compensates for this upfront cost with minimal inference latency during deployment. Because FISOR heavily relies on rejection sampling, significant latency is introduced as every time an action must be selected from N candidates based on cost and reward Q-values. Even when N is set to 1 for FISOR and SafeIFQL and rejection sampling is disabled, inference time remains high due to the multi-step denoising process inherent to both DDPM and Flow Matching policies. This demonstrates that trading a one-time higher training cost for SafeFQL yields a vastly more efficient, highly accurate, single-step policy. Ultimately, this highlights the immense practical effectiveness of SafeFQL for high-frequency, real-time control loops once the model is trained.

\section{Conclusion, Limitations and Future works}
\label{section: conclusions}
We introduced Safe Flow Q-Learning (SafeFQL), a scalable offline safe reinforcement learning framework that synthesizes the expressivity of generative flow models with the rigorous safety principles of Hamilton-Jacobi reachability. By distilling a multi-step flow policy into an efficient one-step actor, SafeFQL eliminates the prohibitive computational costs of iterative action sampling at deployment, achieving an inference time speedup of $2.5 \times$. 
Empirically, SafeFQL maintains competitive rewards while achieving near-zero violations using a single action proposal across both high-dimensional Safety Gymnasium and custom navigation tasks.

While SafeFQL demonstrates robust empirical safety, we identify some scope of algorithmic refinement. Currently we use hard indicator mask for the use of Q-critic functions in Algorithm~\ref{algo:safe-fql} which could theoretically yield a non-smooth loss landscape and might impact training at times. Therefore, one could explore use of continuous masking functions or soft Lagrangian relaxations to improve framework stability, however, that requires hyperparameter finetuning, which can be undesirable.



\bibliography{main}
\bibliographystyle{rlj}

\beginSupplementaryMaterials
\appendix

\section{Description of the Experiments} \label{appendix: env_desc}

\begin{figure}[h]
    \centering
    \includegraphics[width=\linewidth]{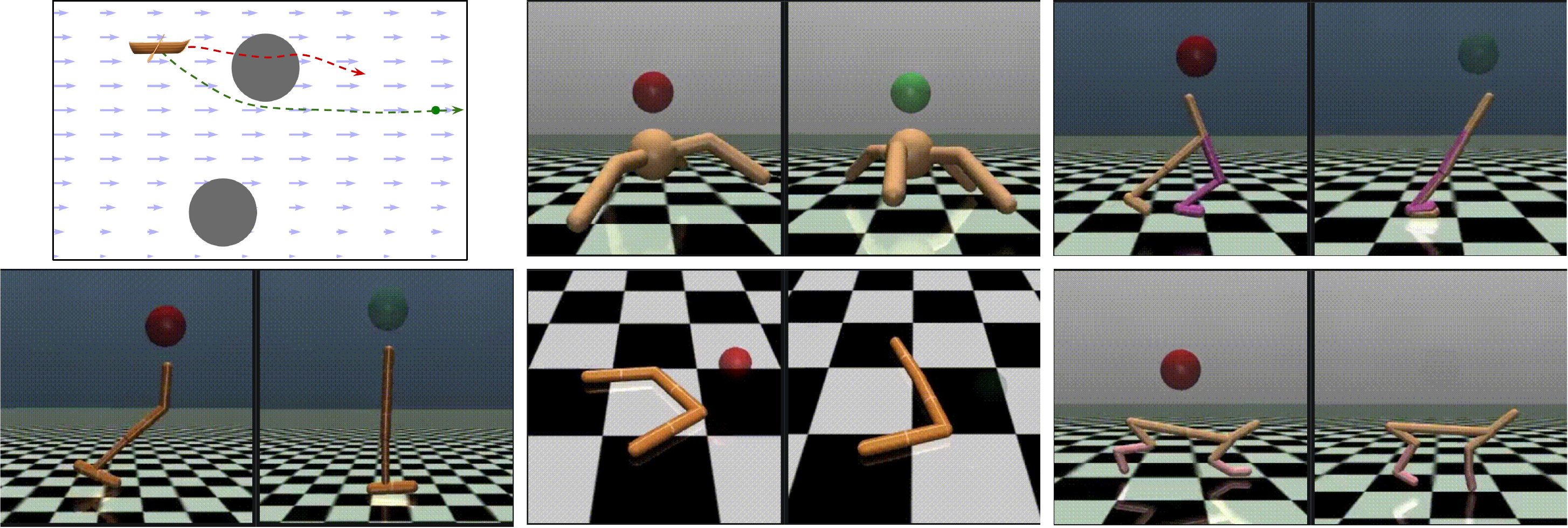}
    \caption{\textbf{Illustration of Evaluation Environments.} \emph{(TopLeft):} Environment depicts the Safe Boat Navigation task with 2 obstacles and a goal point in a drifting river. \emph{(Remaining):} Environments are the standard Safety Gymnasium environments \citep{ji2023safety} from its Safe Velocity suite.}
    \label{fig:placeholder}
\end{figure}

\subsection{Boat Navigation} \label{appendix: Boat2D}
The 2D Boat state is \(x \in \mathcal{X} = [-3,2]\times[-2,2]\), with
\(x = [x_1, x_2]^\top\) representing the boat's Cartesian coordinates \((x_1,x_2)\).
We use a dense, distance-based step reward that encourages progress toward a fixed goal:
\begin{equation}
r(x) \;=\; C \cdot \Big(-\big\|[x_1,x_2]^\top - [x_{g1},x_{g2}]^\top\big\|\Big),
\end{equation}
where the goal is \([x_{g1},x_{g2}]^\top = [0.5,0.0]^\top\) and \(C=0.1\).
Maximizing \(r(x)\) therefore drives the boat toward the goal location.

The discrete-time dynamics are given by
\begin{align*}
x_{1,t+1} &= x_{1,t} + \big(a_{1,t} + 2 - 0.5\,x_{2,t}^2\big)\,\Delta t,\\
x_{2,t+1} &= x_{2,t} + a_{2,t}\,\Delta t,\\
\end{align*}
where \(\Delta t\) is the integration timestep, \((a_{1,t},a_{2,t})\) are the control inputs
satisfying \(a_{1,t}^2 + a_{2,t}^2 \le 1\), and the term \(2 - 0.5 x_{2,t}^2\) models the
state-dependent longitudinal drift along the \(x_1\)-axis.

Obstacles and the failure region are encoded via the safety function \(\ell(x)\).
We define \(\ell(x)\) using the negative of signed distance (plus the obstacle radius) to two circular obstacles:
\begin{equation}
\ell(x) \;:=\; \max\!\big(0.4 - \|x - [-0.5,\,0.5]^\top\|,\; 0.5 - \|x - [-1.0,\,-1.2]^\top\|\big).
\end{equation}
By this definition, \(\ell(x)>0\) indicates that the boat lies inside an obstacle; the super-level set
\(\{x:\ell(x)>0\}\) therefore defines the failure region.

\paragraph{Offline data generation.}
Because this environment is custom, we construct an offline dataset for training and evaluation.
We sample \(2{,}500\) initial states uniformly from \(\mathcal{X}\) and simulate each trajectory for
\(400\) discrete timesteps with step size \(\Delta t = 0.005\) s. During data collection control
inputs are sampled uniformly at random from the admissible action set (subject to the norm constraint),
ensuring a wide variety of state–action coverage for downstream learning of safe controllers.

\subsection{Safety MuJoCo Environments} \label{appendix: SafetyGym}
To evaluate performance on higher-dimensional systems we use the MuJoCo-based Safety Gymnasium
environments \cite{ji2023safety}. These environments implement \emph{safe velocity} tasks in which
the agent incurs a cost whenever its instantaneous velocity exceeds a task-specific threshold.

At each step the environment provides a binary cost defined by the velocity constraint:
\begin{equation}
\text{cost}_t \;=\; \mathbb{I}\big[V_{\text{current},t} > V_{\text{threshold}}\big],
\end{equation}
where \(\mathbb{I}[\cdot]\) is the indicator function. Equivalently, we express safety with the
continuous safety function
\begin{equation}
\ell(x) \;=\; V_{\text{current}}(x) - V_{\text{threshold}},
\end{equation}
so that \(\ell(x)\le 0\) corresponds to the safe set. Using \(\ell(x)\) provides a dense,
continuous safety signal rather than a sparse \(\{0,1\}\) cost.

The per-environment threshold velocities \(V_{\text{threshold}}\), integration timesteps
\(\Delta t\), and action-space bounds \(\mathbb{U}\) are taken from the official documentation;
the values we used are summarized in Table~\ref{tab:velocity_thresh}.

\begin{table}[!h]
\centering
\small
\caption{Velocity thresholds \(V_{\text{threshold}}\), timestep \(\Delta t\), and action-space \(\mathbb{U}\)
for the MuJoCo Safety Gymnasium environments used in our experiments (values from the official docs).}
\begin{tabular}{lccccc}
\toprule
\textbf{Environment} & \textbf{Hopper} & \textbf{HalfCheetah} & \textbf{Ant} & \textbf{Walker2D} & \textbf{Swimmer} \\
\midrule
\(V_{\text{threshold}}\) & 0.7402 & 3.2096 & 2.6222 & 2.3415 & 0.2282 \\
\(\Delta t\) (s) & 0.008 & 0.05 & 0.05 & 0.008 & 0.04 \\
\(\mathbb{U}\) & \([-1,1]^3\) & \([-1,1]^6\) & \([-1,1]^8\) & \([-1,1]^6\) & \([-1,1]^2\) \\
\bottomrule
\end{tabular}
\label{tab:velocity_thresh}
\end{table}

\section{Additional Results} \label{appendix: add_results}

\begin{figure} [h]
  \centering
  \includegraphics[width=0.8\linewidth]{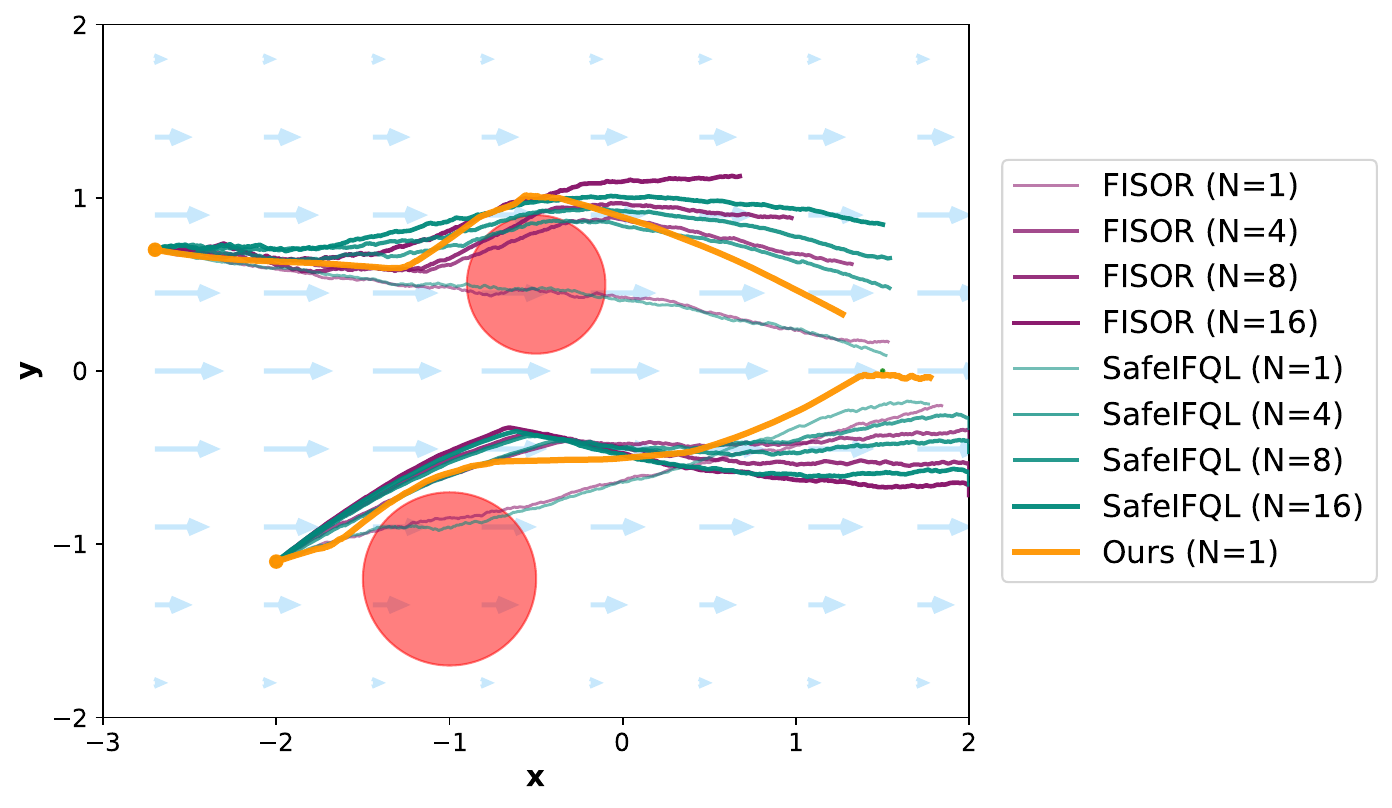}
  \caption{\textbf{Boat Navigation Environment Trajectory Rollouts} for Generative Policy based Frameworks when different candidate action pool sizes (represented by N) are used. For this plot, we use $N \in \{1,4,8,16\}$ for FISOR and SafeIFQL frameworks. For SafeFQL, we stick to N=1 as the framework doesn't require action rejection sampling.}
  \label{fig:boat-trajectories}
\end{figure}

To Visually compare action sample efficiency for the generative policy based methods (FISOR, SafeIFQL and SafeFQL (Ours)), the trajectories across different sample sizes (Figure~\ref{fig:boat-trajectories}) highlights the difference. SafeFQL not only achieves zero collisions across its trajectories, but it also reaches closest to the goal among all safe trajectories by the baselines. Furthermore, because SafeFQL circumvents the need for rejection sampling, it preserves its mathematically optimal nature while completely avoiding the computational hassle of multi-action sampling at inference time.

\section{Experimental Details} \label{appendix: exp_details}

\subsection{Experimental Hardware}
To ensure a fair comparison, all experiments were performed on the same system with a 14th-Gen Intel Core i9-14900KS CPU with 64 GB of RAM and an NVIDIA GeForce RTX 5090 GPU, used for both training and evaluation.

\subsection{Network Architecture and Training Details of the Proposed Algorithm}
We have compiled and listed down all the hyperparameters that we used to perform our experiments and report the results. These training settings for all the  environments are detailed in the Table~\ref{tab:training_details}. For the MuJoCo environments, we use the widely accepted DSRL \cite{liu2024dsrl} dataset. 

\begin{table}[ht]
    \caption{Hyperparameters for the Algorithm (SafeFQL).}
    \centering
    \begin{tabular}{lc}
        \hline
        \textbf{Hyperparameter} & \textbf{Value} \\
        \hline
        Network Architecture & Multi-Layer Perceptron (MLP) \\
        Activation Function & ReLU \\
        Optimizer & Adam optimizer \\
        Learning Rate & $3\times 10^{-4}$ \\
        Discount Factor ($\gamma$) & 0.99 \\
        Time Step Intervals (FM Policy) & 10 \\
        \hline
        \textbf{Boat Navigation} & \\
        \hline
        Number of Hidden Layers ($Q_r, V_r, Q_c, V_c$) & 2 \\
        Number of Hidden Layers ($\pi$) & 3 \\
        Hidden Layer Size (Both) & 256 neurons per layer \\
        Dataset Size & 1M \\
        \hline
        \textbf{Safe Velocity Hopper} & \\
        \hline
        Number of Hidden Layers ($Q_r, V_r, Q_c, V_c$) & 3 \\
        Number of Hidden Layers ($\pi$) & 4 \\
        Hidden Layer Size (Both) & 256 neurons per layer \\
        Dataset Size & 1.32M \\
        \hline
        \textbf{Safe Velocity Half-Cheetah} & \\
        \hline
        Number of Hidden Layers ($Q_r, V_r, Q_c, V_c$) & 3 \\
        Number of Hidden Layers ($\pi$) & 4 \\
        Hidden Layer Size (Both) & 256 neurons per layer \\
        Dataset Size & 249K \\
        \hline
        \textbf{Safe Velocity Ant} & \\
        \hline
        Number of Hidden Layers ($Q_r, V_r, Q_c, V_c$) & 3 \\
        Number of Hidden Layers ($\pi$) & 4 \\
        Hidden Layer Size (Both) & 256 neurons per layer \\
        Dataset Size & 2.09M \\
        \hline
        \textbf{Safe Velocity Walker2D} & \\
        \hline
        Number of Hidden Layers ($Q_r, V_r, Q_c, V_c$) & 3 \\
        Number of Hidden Layers ($\pi$) & 4 \\
        Hidden Layer Size (Both) & 256 neurons per layer \\
        Dataset Size & 2.12M \\
        \hline
        \textbf{Safe Velocity Swimmer} & \\
        \hline
        Number of Hidden Layers ($Q_r, V_r, Q_c, V_c$) & 3 \\
        Number of Hidden Layers ($\pi$) & 4 \\
        Hidden Layer Size (Both) & 256 neurons per layer \\
        Dataset Size & 1.68M \\
        \hline
    \end{tabular}
    \vspace{-1em}
    \label{tab:training_details}
\end{table}

During the training, we set the $\tau$ for expectile regression in section \ref{section: method} to 0.9. And we use clipped double Q-learning \citep{fujimoto2018addressing} for both reward and safety Q-critic functions, taking a minimum of the two Q values. We update the target Q networks using Exponential Moving Average (EMA) where the weight to the new parameters is set to 0.005. Following \citet{kostrikov2022offline}, we clip exponential advantages to $(-\infty, 100]$ in feasible part and $(-\infty, 150]$ in infeasible part. For our paper, we used Flow Q-Learning implementation from \citet{park2025fql}.

\vspace{7em}
\subsection{Hyperparameters for the Baselines} 

Hyperparameters for the remaining safe offline-RL baselines (COptiDICE, BEAR-Lag, CPQ, C2IQL, FISOR) are listed in Table~\ref{tab:baselines_special_nets}. We employ the official implementations of BEAR-Lag, CPQ and COptiDICE from \citet{liu2024dsrl}, C2IQL from \citet{liu2025ciql}, and FISOR from \citet{zheng2024fisor}. For SafeIFQL we follow the hyperparameter choices of \citet{zheng2024fisor} however, Flow-Matching implementation for its IFQL component is adapted from \citet{park2025fql}. All methods are trained on the same DSRL datasets.

\begin{table}[!htbp]
\centering
\small
\caption{Detailed Hyperparameters for Baseline Special Networks. (Layers, Units) notation refers to hidden layers and units per layer. All baselines utilize the DSRL dataset standards \cite{liu2024dsrl}.}
\label{tab:baselines_special_nets}
\begin{tabular}{lll}
\hline
\textbf{Baseline} & \textbf{Network Component} & \textbf{Architecture Specification} \\
\hline
\textbf{C2IQL} \citep{liu2025ciql} & Actor / Critic / Value Nets & MLP, (256, 2) \\
& \textbf{Cost Reconstruction Model} & \textbf{MLP, (5 layers, 512 units each)} \\
& Reward / Cost Advantage & MLP, (256, 2) \\
\hline
\textbf{CPQ} \citep{xu2022constraints} & Actor (Policy) Net & MLP, (256, 2) \\
& \textbf{Constraint-Penalized Q} & \textbf{Ensemble DoubleQCritic, (256, 2)} \\
& Cost Critic Net & MLP, (256, 2) \\
\hline
\textbf{FISOR} \citep{zheng2024fisor} & Diffusion Denoiser (Actor) & DiffusionDenoiserMLP, (256, 2) \\
& \textbf{Feasibility Classifier} & \textbf{MLP, (256, 2), Sigmoid Output} \\
& Energy/Value Guidance & MLP, (256, 2) \\
\hline
\textbf{COptiDICE} \citep{lee2022coptidice} & \textbf{Dual / $\nu$ Network} & \textbf{DualNet (Value-like), (256, 2)} \\
& Actor (Extraction) Net & MLP, (256, 2) \\
\hline
\textbf{BEAR-Lag} (Lag. dual of \citep{kumar2019stabilizing}) & \textbf{VAE (Support Model)} & \textbf{MLP, (750, 2)} \\
& Actor (Policy) Net & SquashedGaussianMLPActor, (256, 2) \\
& Cost / Reward Critics & MLP, (256, 2) \\
\hline
\end{tabular}
\end{table}

\end{document}